\documentclass[11pt,a4paper]{article}
\usepackage[letterpaper]{geometry}
\usepackage{acl2012}
\usepackage{times}
\usepackage{latexsym}
\usepackage{amsmath}
\usepackage{multirow}
\usepackage{graphicx}
\usepackage{comment}
\usepackage{color}
\usepackage{paralist}
\usepackage{booktabs}
\usepackage{amssymb}
\usepackage[round]{natbib}
\usepackage{dblfloatfix}

\usepackage{longtable}
\usepackage[T2A,T1]{fontenc}
\usepackage[utf8]{inputenc}
\usepackage[greek,russian,english]{babel}
\usepackage{pifont}

\usepackage[colorlinks=true]{hyperref}
\hypersetup{allcolors = {black}}%, urlcolor = {blue}}

\newcommand{\textnl}[1]{\textsl{#1}}

\newcommand{\germanic}[1]{\color{blue}{#1}}
\newcommand{\black}[1]{\color{black}{#1}}
\newcommand{\latin}[1]{\color{red}{#1}}

\usepackage[compact]{titlesec}
\title{Native Language Cognate Effects on Second Language  Lexical Choice}

\author{
\fontsize{11}{11}\selectfont{
\begin{tabular}[t]{c@{\extracolsep{4em}}cc}
Ella Rabinovich$^{\star\blacktriangle}$ & Yulia Tsvetkov$^{\dagger}$ & Shuly Wintner$^{\star}$ \\
\end{tabular}
}
\\ \\
\fontsize{11}{11}\selectfont{$^{\star}$Department of Computer Science, University of Haifa} \\
\fontsize{11}{11}\selectfont{$^{\blacktriangle}$IBM Research} \\
\fontsize{11}{11}\selectfont{$^{\dagger}$Language Technologies Institute, Carnegie Mellon University} \\
\fontsize{10.5}{10.5}\selectfont{\tt ellarabi@gmail.com, \tt ytsvetko@cs.cmu.edu \tt shuly@cs.haifa.ac.il} \\
}

\date{}

\begin{document}
\maketitle

\begin{abstract}
We present a computational analysis of cognate effects on the spontaneous linguistic productions of advanced non-native speakers. Introducing a large corpus of highly competent non-native English speakers, and using a set of carefully selected lexical items, we show that the lexical choices of non-natives are affected by cognates in their native language. This effect is so powerful that we are able to reconstruct the phylogenetic language tree of the Indo-European language family solely from the frequencies of specific lexical items in the English of authors with various native languages. We  quantitatively analyze non-native lexical choice, highlighting cognate facilitation as one of the important phenomena shaping the language of non-native speakers.
\end{abstract}

\section{Introduction}
\label{sec:introduction}
Acquisition of vocabulary and semantic knowledge of a second language, including appropriate word choice and awareness of subtle word meaning contours, are recognized as a notoriously hard task, even for advanced non-native speakers. When non-native authors produce utterances in a foreign language (\emph{L2}), these utterances are marked by traces of their native language (\emph{L1}). Such traces are known as \emph{transfer} effects, and they can be phonological (a foreign accent), morphological, lexical, or syntactic. Specifically, psycholinguistic research has shown that the choice of lexical items is influenced by the author's L1, and that non-native speakers tend to choose words that happen to have \emph{cognates} in their native language.

\emph{Cognates} are words in two languages that share both a similar meaning and a similar phonetic (and, sometimes, also orthographic) form, due to a common ancestor in some protolanguage. The definition is sometimes also extended to words that have similar forms and meanings due to \emph{borrowing}. Most studies on cognate facilitation have been conducted with few human subjects, focusing on few words, and the experimental setup was such that participants were asked to produce lexical choices in an artificial setting. We demonstrate that cognates affect lexical choice in L2 spontaneous production on a much larger scale.

Using a new and unique large corpus of non-native English that we introduce as part of this work, we identify a \emph{focus set} of over~1000 words, and show that they are distributed very differently across the ``Englishes'' of authors with various L1s. Importantly, we go to great lengths to guarantee that these words do not reflect specific properties of the various native languages, the cultures associated with them, or the topics that may be relevant for particular geographic regions. Rather, these are ``ordinary'' words, with very little culture-specific weight, that happen to have synonyms in English that may reflect cognates in some L1s, but not all of them. Consequently, they are used differently by authors with different linguistic backgrounds, to the extent that the authors' L1s can be identified through their use of the words in the focus set. The signal of L1 is so powerful, that we are able to reconstruct a linguistic typology tree from the distribution of these words in the Englishes witnessed in the corpus.

We propose a methodology for creating a {focus set} of highly frequent, unbiased words that we expect to be distributed differently across different Englishes simply because they happen to have synonyms with different etymologies, even though they carry very limited cultural weight. Then, we show that simple lexical semantic features (based on the focus set of words) suffice for clustering together English texts authored by speakers of ``closer'' languages; we generate a phylogenetic tree of~31 languages solely by looking at lexical semantic properties of the English spoken by non-native speakers from~31 countries.

The contribution of this work is twofold. First, we introduce the \emph{L2-Reddit corpus}: a large corpus of highly-advanced, fluent, diverse, non-native English, with sentence-level annotations of the native language of each author. Second, we lay out sound empirical foundations for the theoretical hypothesis on cognate effect in L2 of non-native English speakers, highlighting the cognate facilitation phenomenon as one of the important factors shaping the language of non-native speakers.

After discussing related work in Section~\ref{sec:related-work}, we describe the L2-Reddit corpus in Section~\ref{sec:corpus}. Section~\ref{sec:exp} details the methodology we use and our results. We analyze these results in Section~\ref{sec:analysis}, and conclude with suggestions for future research.

\section{Related Work}
\label{sec:related-work}
The language of bilinguals is different. The mutual presence of two linguistic systems in the mind of the bilingual speaker involves a significant cognitive load \citep{schlesinger:2003,hvelplund2014eye,Prior2014,Kroll_Bobb_Hoshino_2014}; this burden is likely to have a bearing on the linguistic productions of the bilingual speaker. Moreover, the presence of more than one linguistic system gives rise to \emph{transfer}: traces of one linguistic system may be observed in the other language \citep{jarvis2008crosslinguistic}. %Psycholinguistic research over the last two decades has demonstrated that the two languages of bilinguals mostly rely on shared neural substrates and cognitive resources \citep{abutalebi2007bilingual,kroll20127,Prior2014}. Current models of the bilingual and learner language systems agree that there is at least some degree of sharing between the languages \citep{french2004understanding,kroll2005models}, and there is wide agreement that conceptual representations are likely shared across the two languages \citep{dunabeitia:etal:2010,francis2005bilingual}.

%These findings are corroborated by research in second language acquisition, which established the unique character of learner language. The entire linguistic system that emerges when second language learners---both children and adults---express meaning in the target language was termed \emph{interlanguage} \citep{Selinker1972, selinker2013rediscovering}. The interlanguage hypothesis assumes the concept of \emph{fossilization}: permanent cessation of language learning \emph{before} the learner has attained the target language norms at all levels of linguistic structure \citep{han:2013}. The fossilization hypothesis predicts that even advanced, fluent non-natives preserve traces that expose their non-nativeness. Some such traces are a result of the influence of the native language of the learner.

Several works addressed the translation choices of bilingual speakers, either within a rich linguistic context (e.g., given a source sentence), or decontextualized. For example,  \citet{de1992determinants} demonstrated that cognate translations are produced more rapidly and accurately than translations that do not exhibit phonetic or orthographic similarity with a source word. This observation was further articulated by \citet{prior2007translation}, who showed that translation choices of L2 speakers were positively correlated with cross-linguistic form overlap of a stimulus word with its target language translations. \citet{prior2011translation} emphasized that ``bilinguals are sensitive to the degree of form overlap between the translation equivalents in the two languages, and show a preference toward producing a cognate translation''. As an example, they showed that the preferred translation of the Spanish \textnl{incidente} to English was \textnl{incident}, and not the alternative translation \textnl{event}, despite the much higher frequency of the latter.

More recent work is consistent with previous research and advances it by highlighting phonologically mediated cross-lingual influences on visual word processing of same- and different-script bilinguals \citep{degani2010semantic,degani2017cross}. Cognate facilitation was also studied using eye tracking \citep{libben2009bilingual,cop2017reading}, demonstrating that the reading of bilinguals is influenced by orthographic similarity of words with their translation equivalents in another language. Crucially, much of this research has been conducted in a laboratory experimental setup; this implies a small number of participants, a small number of target words, and focus on a very limited set of languages. While our research questions are similar, we present a computational analysis of the effects of cognates on L2 productions on a completely different scale: 31 languages, over~1000 words, and thousands of speakers whose spontaneous language production is recorded in a very large corpus.

Corpus-based investigation of non-native language has been a prolific field of recent research. Numerous studies address \emph{syntactic} transfer effects on L2. Such influences from L1 facilitate various computational tasks, including automatic detection of highly competent non-native writers \citep{tomokiyo2001you,bergsma2012stylometric}, identification of the mother tongue of English learners \citep{koppel2005determining,tetreault-blanchard-cahill:2013:BEA,tsvetkov-EtAl:2013:BEA8,malmasi2017report} and typology-driven error prediction in learners' speech \citep{BerzakRK15}. English texts produced by native speakers of a variety of languages have been used to reconstruct phylogenetic trees, with varying degrees of success \citep{NagataW13,BerzakRK14}. Syntactic preferences of professional translators were exploited to reconstruct the Indo-European language tree \citep{rabinovich-ordan-wintner:2017:Long}. Our  study is also corpus-based; but it stands out as it focuses not on the distribution of function words or (shallow) syntactic structures, but rather on the unique use of cognates in L2.

From the \emph{lexical} perspective, L2 writers have been shown to produce more overgeneralizations, use more frequent words and words with a lower degree of ambiguity \citep{hinkel2002second,Crossley2011}. Several studies addressed cross-linguistic influences on semantic acquisition in L2, investigating the distribution of collocations \citep{siyanova2015collocation,kochmar2017modelling} and formulaic language \citep{paquot2012formulaic} in learner corpora.
We, in contrast, address highly-fluent, advanced non-natives in this work.

\citet{nastase2017word} presented the first attempt to leverage etymological information for the task of native language identification of English learners. They sowed the seeds for exploitation of etymological clues in the study of non-native language, but their results were very inconclusive.

In contrast to the learner corpora that dominate studies in this field \citep{icle,efcamdat,toefl}, our corpus contains spontaneous productions of advanced, highly proficient non-native speakers, spanning over 80K topical threads, by 45K distinct users from 50 countries (with 46 native languages). To the best of our knowledge, this is the first attempt to computationally study the effect of L1 cognates on L2 lexical choice in productions of competent non-native English speakers, certainly at such a large scale.

\section{The \emph{L2-Reddit corpus}}
\label{sec:corpus}
One contribution of this work is the collection, organization and annotation of a large corpus of highly-fluent non-native English. We describe this new and unique corpus in this section.

\subsection{Corpus mining}
Reddit\footnote{\url{https://www.reddit.com/}} is an online community-driven platform consisting of numerous forums for news aggregation, content rating, and discussions. As of $2017$, it has over $200$ million unique users, ranking the fourth most visited website in the US. Content entries are organized by areas of interest called \emph{subreddits},\footnote{Subreddits are typically denoted with a leading r/, for example r/linguistics is the `linguistics' subreddit.} ranging from main forums that receive much attention to smaller ones that foster discussion on niche areas. Subreddit topics include news, science, movies, books, music, fitness and many others.

\paragraph{Collection of author metadata} We collected a large dataset of posts (both initial submissions and subsequent comments) using an API especially designed for providing search capabilities on Reddit content.\footnote{\url{https://github.com/pushshift/api}} We focused on several subreddits (r/Europe, r/AskEurope, r/EuropeanCulture, r/EuropeanFederalists, r/Eurosceptics) whose content is generated by users who specified their country as a \emph{flair} (metadata attribute). Although categorized as `European', these subreddits are used by people from all over the world, expressing views on politics, legislation, economics, culture, etc.

In the absence of a restrictive policy, multiple flair alternatives often exist for the same country, e.g., `CROA' and `Croatia' for Croatia. Additionally, distinct flairs are sometimes used for regions, cities, or states of big European countries, e.g., `Bavaria' for Germany. We (manually) grouped flairs representing the same country into a single cluster, reducing 489 distinct flairs into 50 countries, from Albania to Vietnam. The posts in the Europe-related subreddits constitute our \emph{seed corpus}, comprising 9M sentences (160M tokens) by over 45K distinct users.

\paragraph{Dataset expansion} A typical user activity in Reddit is not limited to a single thread, but rather spreads across multiple, not necessarily related, areas of interest. Once the authors' country is determined based on their European submissions, their entire Reddit footprint can be associated with their profile, and, therefore, with their country of origin. We extended our seed corpus by mining \emph{all} submissions of users whose country flair is known, querying all Reddit data spanning years 2005-2017. The final dataset thus contains over 250M sentences (3.8B tokens) of native and non-native English speakers, where each sentence is annotated with its author's country of origin. The data covers posts by over 45K authors and spans over 80K subreddits.\footnote{The annotated dataset is freely available at \url{http://cl.haifa.ac.il/projects/L2}. To protect the anonymity of Reddit users, the released dataset does not expose any author identifying information.}

\paragraph{Focus on ``large'' languages} For the sake of robustness, we limited the scope of this work to (countries whose L1s are) the Indo-European (IE) languages; and only to those countries whose users had at least 500K sentences in the corpus. Additionally, we excluded multilingual countries, such as Belgium and Switzerland. Consequently, the final set of Reddit authors considered in this work originate from~31 countries, which represent the three main IE language families: \emph{Germanic} (Austria, Denmark, Germany, Iceland, Netherlands, Norway, Sweden); \emph{Romance} (France, Italy, Mexico, Portugal, Romania, Spain); and \emph{Balto-Slavic} (Bosnia, Bulgaria, Croatia, Czech, Latvia, Lithuania, Poland, Russia, Serbia, Slovakia, Slovenia, Ukraine). In addition, we have data authored by native English speakers from Australia, Canada, Ireland, New Zealand, the UK and the US.

\paragraph {Correlation of country annotation with L1} We view the country information as an accurate, albeit not perfect, proxy for the native language of the author.\footnote{We therefore use the terms `user country', `native language' and `L1' interchangeably henceforth.} We acknowledge that the L1 information is noisy and may occasionally be inaccurate. We therefore evaluated the correlation of the country flair with L1 by means of supervised classification: our assumption is that if we can accurately distinguish among users from various countries using features that reflect language, rather than culture or content, then such a correlation indeed exists.

We assume that the native language of speakers ``shines through'' mainly in their syntactic choices. Consequently, we opted for (shallow) syntactic structures, realized by function words (FW) and n-grams of part-of-speech (POS) tags, rather than geographical and topical markers, that are reflected best by content words. Aiming to disentangle the effect of native language we randomly shuffled texts produced by all authors from each country, thereby ``blurring out'' any topical (i.e., subreddit-specific) or authorial trace. Consequently, we assume that the separability of texts by country can be attributed to the only distinguishing linguistic variable left: the dimension of the native language of a speaker.

We classified 200 chunks of randomly sampled 100 sentences form each country into
\begin{inparaenum}[(i)]
\item native vs.\ non-native English speakers,
\item the three IE language families, and
\item 45 individual L1s, where the six English-speaking countries are unified under the native-English umbrella.
\end{inparaenum}
Using over 400 function words and top-300 most frequent POS-trigrams, we obtained 10-fold cross-validation accuracy of 90.8\%, 82.5\% and 60.8\%, for the three scenarios, respectively. We conclude, therefore, that the country flair can be viewed as a plausible proxy for the native language of Reddit authors.

\paragraph{Initial preprocessing}
\label{sec:preprocessing}
Several preprocessing steps were applied on the dataset. We
\begin{inparaenum}[(i)]
\item removed text by users who changed their country flair within their period of activity;
\item excluded non-English sentences;\footnote{We used the \emph{polyglot} language detection tool (\url{http://polyglot.readthedocs.io}).} and
\item eliminated sentences containing single non-alphabetic tokens.
\end{inparaenum}
The final corpus comprises over 230M sentences and 3.5B tokens.

\subsection{Evaluation of author proficiency}
Unlike most corpora of non-native speakers, which focus on \emph{learners} (e.g., ICLE \citep{icle}, EFCAMDAT \citep{efcamdat}, or the TOEFL dataset  \citep{toefl}), our corpus is unique in that  it is composed by fluent, advanced non-native speakers of English.
We verified that, on average, Reddit users possess excellent, near-native command of English by comparing three distinct populations:
\begin{inparaenum}[(i)]
\item Reddit native English authors, defined as those tagged for one of the English-speaking countries: Australia, Canada, Ireland, New Zealand, and the UK. We excluded texts produced by US authors due to the high ratio of the US immigrant population;
\item Reddit non-native English authors; and
\item A population of English learners, using the TOEFL dataset \citep{toefl}; here, the proficiency of authors is classified as low, intermediate, or high.
\end{inparaenum}

We compared these populations across various indices, assessing their proficiency with several commonly accepted lexical and syntactic complexity measures \citep{LU201516,kyle2015automatically}. Lexical richness was evaluated through type-to-token ratio (TTR), average age-of-acquisition (in years) of lexical items
\citep{Kuperman2012}, and mean word rank, where the rank was retrieved from a list of the entire Reddit dataset vocabulary, sorted by word frequency in the corpus. Syntactic complexity was assessed using mean length of T-units (TU; the minimal terminable unit of language that can be considered a grammatical sentence), and the ratio of complex T-units (those containing a dependent clause) to all T-units in a sentence.

\begin{table*}[h!]
\centering
\begin{tabular}{lccccc}
\multicolumn{1}{c}{\textbf{Population}} & \multicolumn{1}{c}{\textbf{Mean TU length}} & \multicolumn{1}{c}{\textbf{Complex TU ratio}} & \multicolumn{1}{c}{\textbf{TTR}} & \multicolumn{1}{c}{\textbf{Mean word rank}} & \multicolumn{1}{c}{\textbf{AoA}} \\ \hline
Learners (low) & 15.583 & 0.513 & 0.089 & 1172.19 & 5.186 \\
Learners (medium) & 16.357 & 0.534 & 0.106 & 1504.01 & 5.317 \\
Learners (high) & 17.468 & 0.528 & 0.124 & 1852.64 & 5.562 \\ %\hline
\textit{Reddit non-natives} & \textit{19.528} & \textit{0.633} & \textit{0.174} & \textit{1960.62} & \textit{5.524} \\
Reddit natives & 20.154 & 0.658 & 0.179 & 2063.89 & 5.575 \\
\end{tabular}
\caption{Evaluation of the English proficiency of non-native Reddit users.}
\label{tbl:proficiency}
\end{table*}

Table~\ref{tbl:proficiency} reports the results. Across almost all indices, the level of Reddit non-natives is much higher than even the advanced TOEFL learners, and almost on par with Reddit natives.

\section{L1 cognate effects on L2 lexical choice}
\label{sec:exp}
\subsection{Hypotheses}
\label{sec:hypotheses}
Cognates are words in two languages that share both a similar meaning and a similar form. Our main hypothesis is that non-native speakers, when required to pick an English word that has a set of synonyms, are more likely to select a lexical item that has a cognate in their L1. We therefore expect the effect of L1 cognates to be reflected in the frequency of their English counterparts in the spontaneous productions of L2 speakers. Moreover, we expect similar effects, perhaps to a lesser extent, in the contextual usage of certain words, reflecting collocations and subtle contours of word meanings that are transferred from L1. The different contexts that certain words are embedded in (in the Englishes of speakers with different L1 backgrounds) can be captured by means of distributional semantics.

Furthermore, we hypothesize that the effect of L1 is powerful to an extent that facilitates clustering of Englishes produced by non-natives with ``similar'' L1s; specifically, L1s that belong to the same language family. ``Similar'' L1s may reflect both typological and areal closeness: for example, we expect the English spoken by Romanians to be similar both to the English of Italians (as both are Romance languages) and to the English of Bulgarians (as both are Balkan languages). Ultimately, we aim to reconstruct the IE language phylogeny, reflecting historical and areal evolution of the subsets of Germanic, Romance and Balto-Slavic languages over thousands of years, from non-native English only.

While lexical transfer from L1 is a known phenomenon in \emph{learner} language, we hypothesize that its signal is present also in the language of highly competent non-native speakers. Mastering the nuances of lexical choice, including subtle contours of word meaning and the correct context in which words tend to occur, are key factors in advanced language competence. %Nevertheless, we hypothesize that L1 cognates affect the spontaneous productions of even advanced, highly proficient non-native English speakers.
The L2-Reddit corpus provides a perfect environment for testing this hypothesis.

\subsection{Selection of a focus set of words}
\label{sec:focus-set}
Our goal is to investigate non-native speakers' choice of lexical items in English. We address this task by defining a set of English words that have at least one synonym; ideally, we would like the various synonyms to have different etymologies, and in particular, to have different cognates in different language families. English happens to be a particularly good choice for this task, since in spite of its Germanic origins, much of its vocabulary evolved from Romance, as a great number of words were borrowed from Old French during the Norman occupation of Britain in the 11th century.

To trace the etymological history of English words we used Etymological Wordnet (EW), a database that contains information about the ancestors of over 100K English words, about 25K of them in contemporary English \citep{deMeloEtymWN2014}. For each word recorded in EW, the full path to its root can be reconstructed. Intuitively, an English word with Latin roots may exhibit higher (phonetic and orthographic) proximity to its Romance languages' counterparts. Conversely, an English word with a Proto-Germanic ancestor may better resemble its equivalents in Germanic languages.

We selected from EW all the nouns, verbs, and adjectives. For each such word \textit{w}, we identified the synset of \textit{w} in WordNet, choosing only the first (i.e., most prominent) sense of \textit{w} (and, in particular, corresponding to the most frequent part-of-speech (POS) category of \textit{w} in the L2-Reddit dataset). Then, we retained only those words that had  synonyms, and only those whose synonyms had at least two different etymological paths, i.e., synonyms rooted in different ancestors. For example, we retained the synset \{\textnl{heaven}, \textnl{paradise}\}, since the former is derived from Proto-Germanic \textnl{*himin-}~, while the latter is derived from Greek \selectlanguage{greek}\textnl{παράδεισος}\selectlanguage{english} (via Latin and Old French).

%Aiming to create multiple sets of synonyms with words bearing close meaning but different etymological roots, we retrieved for each \textit{noun}, \textit{verb} or \textit{adjective} word \textit{w} in etymological database its first WordNet synset\footnote{When querying WordNet for word synonyms, a sense with the most frequent POS tag in the collection was considered.}, further considering all terms with full etymological path that differs from that of \textit{w}. Focusing on widespread, ubiquitous terms, we further filtered out all words with less than 100 occurrences in the corpus. Words with unknown etymological roots (i.e., not recorded in the etymological database) were excluded from the analysis.

Furthermore, to capture the bias of non-native speakers toward their L1 cognate, it makes sense to focus on a set of easily interchangeable synonyms, e.g., \{\textnl{divide}, \textnl{split}\}. In contrast, consider an unbalanced synset \{{\textnl{kiss, buss, osculation}}\}: presumably, the prevalent alternative \textnl{kiss} is likely to be used by all speakers, regardless of their native language. To eliminate such cases, we excluded synsets that were dominated by a single alternative (with a frequency of over 90\% in our corpus), compared to other synonymous choices. Table \ref{tbl:etymology} illustrates a few examples of synonym sets with their etymological origins.

\begin{table*}[hbt]
\centering
\begin{tabular}{ll}
\multicolumn{1}{c}{\textbf{Synonym set}} & \multicolumn{1}{c}{\textbf{Etymological path to root}} \\ \hline
\textnl{cargo} (N) & Spanish: \textnl{cargo} $\leftarrow$ Spanish: \textnl{cargar} $\leftarrow$ Late Latin: \textnl{carricare} \\
\textnl{freight} (N) & Mid.\ English: \textnl{freyght} $\leftarrow$ Mid.\ Low German: \textnl{vrecht} $\leftarrow$ Proto-Germanic \textnl{*fra-} + \textnl{*aihtiz}\\ \hline
\textnl{weary} (Adj) & Mid.\ English: \textnl{wery} $\leftarrow$ Old English: \textnl{w\=eri\.g} $\leftarrow$ Proto-Germanic: \textnl{*w\=or\={\i}gaz} \\
\textnl{fatigue} (Adj) & French: \textnl{fatigue} $\leftarrow$ French: \textnl{fatiguer} $\leftarrow$ Latin: \textnl{fatigare} \\ \hline
\textnl{exaggerate} (V) & Latin: \textnl{exaggerare} $\leftarrow$ Latin: \textnl{ex-} + Latin: \textnl{aggerare} \\
\textnl{overdo} (V) & English: \textnl{over} + \textnl{do} \\
\end{tabular}
\caption{Etymological roots of example synonym sets with corresponding part-of-speech.}
\label{tbl:etymology}
\end{table*}

\paragraph{Eliminating cultural bias}
Although our Reddit corpus spans over 80K topical threads and 45K users, posts produced by authors from neighboring countries may carry over markers with similar geographical or cultural flavor. For example, we may expect to encounter \textnl{soviet} more frequently in posts by Russians and Ukrainians, \textnl{wine} in texts of French or Italian authors, and \textnl{refugees} in posts by German users. While they may be typical to a certain population group, such terms are totally unrelated to the phenomenon we address here, and we therefore wish to eliminate them from the focus set of words.

A common way to identify elements that are statistically over-represented in a particular population, compared to another, is \emph{log-odds ratio informative Dirichlet prior} \citep{monroe2008fightin}. We employed this approach to discover words that were overused by authors of a certain country, where posts from each country (a category under test) were compared to all the others (the background). We used the strict log-odds score of $-5$ as a threshold for filtering out terms associated with a certain country.\footnote{The threshold was set by preliminary experiments, without any further tuning.} Among the terms eliminated by this procedure were \textnl{genocide} for Armenia, \textnl{hockey} for Canada and \textnl{independence} for the UK. The final focus set of words thus consists of neutral, ubiquitous sets of synonyms, varying in their etymological roots. It comprises~540 synonym sets and~1143 distinct words.

\subsection{Model}
We hypothesize (Section~\ref{sec:hypotheses}) that L1 effects on lexical choice are so powerful, even with advanced non-native speakers, that it is possible to reconstruct the IE language phylogeny, reflecting historical and areal evolution over thousands of years, from non-native English only. We now describe a simple yet effective framework for clustering the Englishes of authors with different L1s, integrating both word frequencies and semantic word representations of the words in our focus set (Section~\ref{sec:focus-set}).

\subsubsection{Data cleanup and abstraction}
Aiming to learn word representations for the lexical items in our focus set, we want the contextual information to be as free as possible from strong geographical and cultural cues. We therefore process the corpus further. First, we identified named entities (NEs) and systematically replaced them by their type. We used the implementation available in the \emph{spacy} Python package,\footnote{\url{https://spacy.io}} which supports a wide range of entities (e.g., names of people, nationalities, countries, products, events, book titles, etc.), at state-of-the-art accuracy. Like other web-based user generated content, the Reddit corpus does not adhere to strict casing rules, which has detrimental effects on the accuracy of NE identification. To improve the tagging accuracy, we applied a preprocessing step of `truecasing', where each {token} $w$ was assigned the case (lower, upper, or upper-initial) that maximized the likelihood of the consecutive tri-gram $\langle w_{pre}$, $w$, $w_{post} \rangle$ in the Corpus of Contemporary American English (COCA).\footnote{\url{https://www.ngrams.info}}
For example, the tri-gram `the us people'  was converted to `the US people', but `let us know' remained unchanged. When a tri-gram was not found in the COCA n-gram corpus, we employed fallback to unigram probability estimation. Additionally, we replaced all non-English words with the token `UNK'; and all web links, subreddit (e.g., r/compling) and user (u/userid) pointers  with the `URL' token.\footnote{The cleaned, abstracted subset of the corpus is also available at \url{http://cl.haifa.ac.il/projects/L2}. The cleanup code is available at \url{https://github.com/ellarabi/reddit-l2}.}

%To overcome this issue, we automatically corrected the case of each token $w$ in the corpus in the following way. We computed the frequencies of trigrams consisting of one token to the left and to the right of the target token $w$, where $w$ can either be all lower case, all upper case or lower case with an initial capital letter. The frequencies were computed from the Corpus of Contemporary American English.\footnote{\url{https://www.ngrams.info}} Then, we replaced $w$ with its most likely form (falling back to unigram frequencies when the trigram frequency was zero). So, for example, \textnl{the us people} was converted to \textnl{the US people}, but \textnl{let us know} remained unchanged.

\subsubsection{Distance estimation and clustering}
\label{sec:distance-estimation}
\citet{bamman-dyer-smith:2014} introduced a model for incorporating contextual information (such as geography) in learning vector representations. They proposed a joint model for learning word representations in situated language, a model that ``includes information about a subject (i.e., the speaker), allowing to learn the contours of a word's meaning that are shaped by the context in which it is uttered''. Using a large corpus of tweets, their joint model learned word representations that were sensitive to geographical factors, demonstrating that the usage of \textnl{wicked} in the United States (meaning \textnl{bad} or \textnl{evil}) differs from that in New England, where it is used as an adverbial intensifier (\textnl{my boy's wicked smart}).

We leveraged this model to uncover linguistic variation grounded in the different L1 backgrounds of non-native Reddit speakers. We used equal-sized random samples of 500K sentences from each country to train a model of vector representations. The model comprises representation of every vocabulary item in each of the 31 Englishes; e.g., 31 vectors are generated for the word \textnl{fatigue}, presumably reflecting the subtle divergences of word semantics, rooted in the various L1 backgrounds of the authors.

%\paragraph{Building countries pairwise distance matrix}
In order to cluster together Englishes of speakers with ``similar'' L1s, we need a measure of distance between two English texts. This measure is based on two constituents: word frequencies and word embeddings. Given two English texts originating from different countries, we computed for each word $w$ in our focus set
\begin{inparaenum}[(i)]
\item the difference in the frequency of $w$ in the two texts; and
\item the distance between the vector representations of $w$ in these texts, estimated by cosine similarity of the two corresponding word vectors.
\end{inparaenum}
We employed the popular \emph{weighted product model} to integrate the two arguments. The word vector component was assigned a higher weight as the frequency of $w$ in the collection increases; this is motivated by the intuition that learning the semantic relationships of a word benefits from vast usage examples. We therefore weigh the embedding constituent proportionally to the word's frequency in the dataset, and assign the complementary weight to the difference of frequencies.

Formally, given two English texts $E_{L_i}$ and $E_{L_j}$, with $L_i$ and $L_j$ native languages, and given a word $w$ in the focus set, let $f_i$ and $f_j$ denote the frequencies of $w$ in $E_{L_i}$ and $E_{L_j}$, respectively. Let $p_w$ be the probability of $w$ in the entire collection. We further denote the vector space representation of $w$ in $E_{L_i}$ by $v_i$, and the representation of $w$ in $E_{L_j}$ by $v_j$. Then, the distance between $E_{L_i}$ and $E_{L_j}$ with respect to the word $w$ is:

\begin{equation}
D_{ij}(w) = (|f_i - f_j|)^{1 - p_w} \times (1 - cos(v_i, v_j))^{p_w}
\label{eq-distance-word}
\end{equation}

\noindent
The final distance between $E_{L_i}$ and $E_{L_j}$ is given by averaging $D_{ij}$ over all words in the focus set $FS$:

\begin{equation*}
D_{ij} = \dfrac{(\sum_{w \in FS}D_{ij}(w))}{|FS|}
\label{eq-distance}
\end{equation*}

Finally, we constructed a symmetric distance matrix (31 $\times$ 31) $M$ by setting $M[i,j]=D_{ij}$. We used Ward's hierarchical clustering\footnote{\url{https://docs.scipy.org/doc/scipy/reference/generated/scipy.cluster.hierarchy.linkage.html}} with the Euclidean distance metric to derive a tree from the distance matrix M.

We considered several other weighting alternatives, including assignment of constant weights to the two factors in Equation~\ref{eq-distance-word}; they all resulted in inferior outcomes. We also corroborated the relative contribution of the two components by using each of them alone. While considering only frequencies resulted in a slightly inferior outcome (see Section~\ref{sec:evaluation}), using word representations alone produced a completely arbitrary result.

\subsection{Results}
\label{sec:clustering-results}
\begin{figure}[hbt]
\begin{center}
\includegraphics[width=\columnwidth]{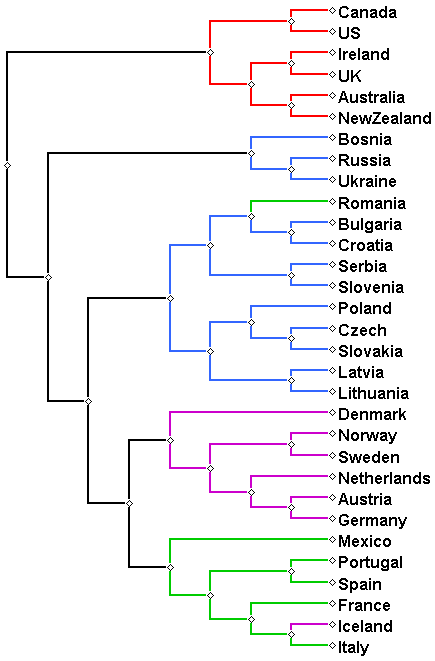}
\end{center}
\caption{Language typology reconstructed from non-native Englishes using features reflecting lexical choice. Countries that belong to the same phylogenetic family (according to the gold tree) share identical color. E.g., Iceland is colored purple, like other Germanic languages, even though it is assigned to the Romance cluster.}
\label{fig:typology}
\end{figure}

The resulting tree is depicted in Figure~\ref{fig:typology}. The reconstructed language typology reveals several interesting observations. First, and much expectedly, all native English speakers are grouped together into a single, distant sub-tree, implying that similarities exhibited by the lexical choices of native speakers go beyond geographical and cultural differences. The Englishes of non-native speakers are clustered into three main language families: Germanic, Romance, and Balto-Slavic. Notably, Spanish-speaking Mexico is clustered with its Romance counterparts. The firm Balto-Slavic cluster reveals historical relations between languages by generating coherent sub-branches: the Czech Republic and Slovakia, Latvia and Lithuania, as well as the relative proximity of Serbia and Croatia. In fact, former Yugoslavia is clustered together, except for Bosnia, which is somewhat detached. Similar close ties can be seen between Austria and Germany, and between Portugal and Spain.

Another interesting phenomenon is captured by English texts of authors from Romania: their language is assigned to the Balto-Slavic family, implying that the deep-rooted areal and cultural Balkan influences left their traces in the Romanian language, which in turn, is reflected in the English productions of native Romanian authors. Unfortunately, we cannot explain the location of Iceland.

\begin{figure*}[hbt]
\begin{center}
\includegraphics[height=5cm, width=16.5cm]{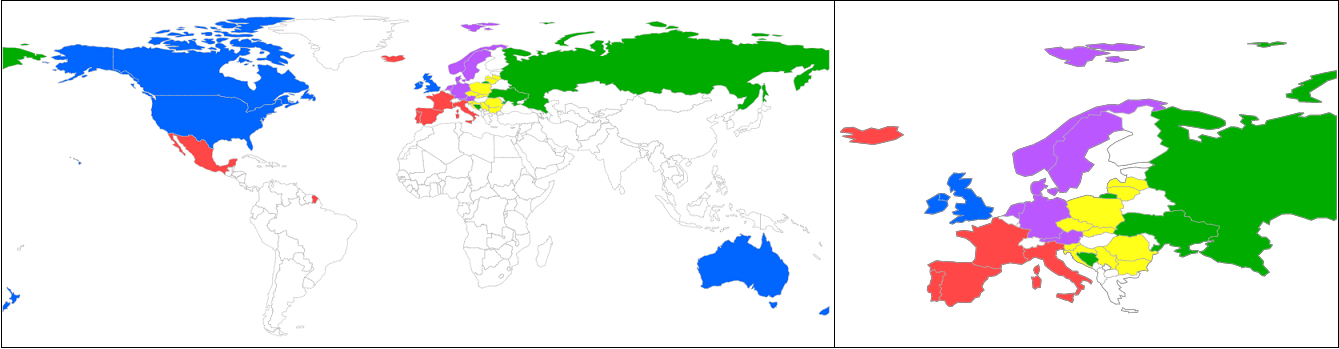}
\end{center}
\caption{Countries by clusters: World (on the left) and Europe (on the right) views. Countries assigned to the same flat cluster by the \textnl{clustering procedure} (Section~\ref{sec:clustering-results}) share identical color, e.g., the wrongly assigned Iceland shares the red color with the Romance-language speaking countries. Countries not included in this work are uncolored.}
\label{fig:clusters-world-europe}
\end{figure*}

A geographical view mirroring the language phylogeny is presented in Figure~\ref{fig:clusters-world-europe}. Flat clusters were obtained from the hierarchy using the \emph{scipy fcluster} method\footnote{\url{https://docs.scipy.org/doc/scipy/reference/generated/scipy.cluster.hierarchy.fcluster.html}}  with defaults.

\begin{figure}[hbt]
\begin{center}
\includegraphics[width=\columnwidth]{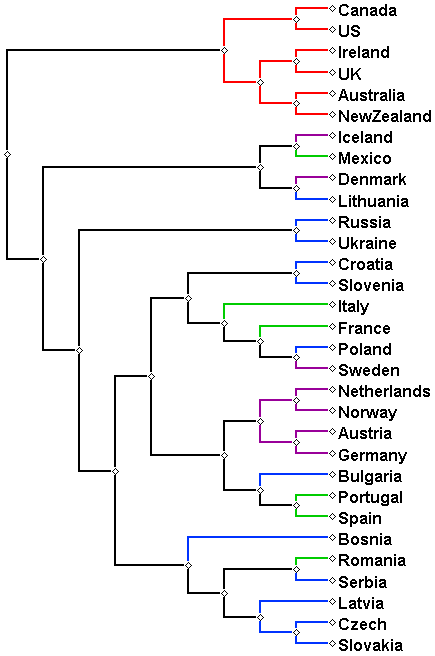}
\end{center}
\caption{Language typology reconstructed from a randomly selected focus set of 1143 words.}
\label{fig:random-typology}
\end{figure}

This outcome, obtained using only lexical semantic properties (word frequencies and word embeddings) of English authored by various non-native speakers, is a strong indication of the power of L1 influence on L2 speakers, even highly fluent ones. These results are strongly dependent on the choice of focus words: we carefully selected words that on one hand lack any cultural or geographical bias toward one group of non-natives, but on the other hand have synonyms with different etymologies. As an additional validation step, we generated a language tree using exactly the same methodology but a different set of focus words. We randomly sampled 1143 words from the corpus, controlling for country-specific bias but \emph{not} for the existence of synonyms with different etymologies. Although some of the intra-family ties were captured (in particular, all native speakers were clustered together), the resulting tree (Figure~\ref{fig:random-typology}) is far inferior.

We also conducted an additional experiment, including multilingual Belgium and Switzerland in the set of countries. While the L1 of speakers cannot be determined for these two countries, presumably Belgium is dominated by Dutch and French, and Switzerland by German and French. Indeed, both countries were assigned into the Germanic language family in our clustering experiments.

\subsection{Evaluation}
\label{sec:evaluation}
To better assess the quality of the reconstructed trees we now provide a quantitative evaluation of the language typologies obtained by the various experiments. We adopt the evaluation approach of \citet{rabinovich-ordan-wintner:2017:Long}, who introduced a distance metric between two trees, defined as the sum of the square differences between all leaf-pair distances in the two trees. More specifically, given a tree of $N$ leaves, $l_i$, $i\in [1..N]$, the distance between two leaves $l_i$, $l_j$ in a tree $\tau$, denoted $D_{\tau}(l_i,l_j)$, is defined as the length of the shortest path between $l_i$ and $l_j$. The distance $Dist(\tau,g)$ between a generated tree $\tau$ and the gold tree $g$ is then calculated by summing the square differences between all leaf-pair distances in the two trees:

\begin{equation*}
Dist(\tau,g) = {\sum_{i,j \in [1..N]; i \neq j}(D_{\tau}(l_i,l_j)-D_g(l_i,l_j))^2}
\end{equation*}

We used the Indo-European tree in \emph{Glottolog}\footnote{\url{http://glottolog.org/}} as our gold standard, pruning it to contain the set of 31 languages considered in this work. For the sake of comparison, we also present the distance obtained for a completely random tree, generated by sampling a random distance matrix from the uniform (0, 1) distribution. The reported random tree evaluation score is averaged over 100 experiments.

Table~\ref{tbl:evaluation} presents the results. All distances are normalized to a zero-one scale, where the bounds, zero and one, represent the identical and the most distant tree with respect to the gold standard, respectively. Much expectedly, the random tree is the worst one, followed closely by the tree reconstructed from a random sample of over 1000 words sampled from the corpus (Figure~\ref{fig:random-typology}). The best result is obtained by considering both word frequencies and representations, being only slightly superior to the tree reconstructed using word frequencies alone. The latter result corroborates the aforementioned observation (Section~\ref{sec:distance-estimation}) and further posits word frequencies as the major factor affecting the shape of the obtained phylogeny.

\begin{table}[hbt]
\centering
%\resizebox{\linewidth}{!}{
\begin{tabular}{lc}
\multicolumn{1}{c}{\textbf{Features used}} & \multicolumn{1}{c}{\textbf{Distance}} \\ \hline
Random tree & 1.000 \\
Randomly sampled words (Figure~\ref{fig:random-typology}) & 0.857 \\
Focus set with frequencies only & 0.497 \\
\ + embeddings (Figure \ref{fig:typology}) & \textbf{0.469} \\
\end{tabular}
%}
\caption{Normalized distance between a reconstructed and the gold tree; lower distances indicate better result.}
\label{tbl:evaluation}
\end{table}

\section{Analysis}
\label{sec:analysis}
The results described in Section~\ref{sec:clustering-results} empirically support the intuition that cognates are one of the factors that shape lexical choice in productions of non-native authors. In this section we perform a closer analysis of the data, aiming to capture the subtle yet systematic distortions that help distinguish between English texts of speakers with different L1s.

\paragraph{Quantitative analysis} Given a synonym set $s \in FS$, consisting of words $\langle w_1, w_2, ..., w_n \rangle$, and two English texts with two different L1s, $E_{L_i}$ and $E_{L_j}$, we computed the counts of the synset words in these texts, and further normalized the counts by the total sum, yielding probabilities. We denote the probability distribution of a synset $s=\langle w_1, w_2, ..., w_n \rangle$ in $E_{L_i}$ by

\begin{equation*}
P^s_i=\langle p_i(w_1), p_i(w_2), ..., p_i(w_n) \rangle
\end{equation*}

\noindent
The different usage patterns of a synonym set $s$ across two Englishes can then be estimated using the Jensen-Shannon divergence (JSD) between the two probability distributions:

\begin{equation}
\label{eq:divergence}
div_{ij}(s) = JSD(P^s_i, P^s_j)
\end{equation}

\noindent
We expect that ``close'' L1s will have lower divergence, whereas L1s from different language families will exhibit higher divergences.

Table~\ref{tbl:divergence-analysis} presents the top twenty synonym sets for the arbitrarily chosen Germany--Spain country pair, ranked by divergence (Equation~\ref{eq:divergence}). The overuse of \textnl{hinder} by German authors may be attributed to its German \textnl{behindern} cognate, whereas Spanish users' preference of \textnl{impede} is probably attributable to its Spanish \textnl{impedir} equivalent. A Spanish cognate for \textnl{plantation}, \textnl{plantaci\'{o}n}, possibly explains the clear preference of Spanish native speakers for this alternative, compared to the more popular choice of German authors, \textnl{grove}, which has Germanic etymological origins.

\begin{table*}[hbt]
\centering
\begin{tabular}{lll}
\multicolumn{1}{c}{\textbf{Synonym set $s$}} & \multicolumn{1}{c}{\textbf{$P_{\textit{Germany}}^{s}$}} & \multicolumn{1}{c}{\textbf{$P_{\textit{Spain}}^{s}$}} \\ \hline
$\langle${\germanic{hinder} \latin{impede}}$\rangle$ & (0.909, 0.091) & (0.69, 0.31) \\
$\langle${\germanic{grove} \black{orchard} \latin{plantation}}$\rangle$ & (0.643, 0.214, 0.143) & (0.227, 0.068, 0.705) \\
$\langle${\germanic{weariness} \black{tiredness} \latin{fatigue}}$\rangle$ & (0.167, 0.208, 0.625) & (0.017, 0.119, 0.864) \\
$\langle${\germanic{yarn} \black{recital} \latin{narration}}$\rangle$ & (0.55, 0.1, 0.35) & (0.22, 0.15, 0.63) \\
$\langle${\germanic{bloom} \germanic{blossom} \latin{flower}}$\rangle$ & (0.25, 0.143, 0.607) & (0.085, 0.098, 0.817) \\
$\langle${\latin{conceivable} \latin{imaginable}}$\rangle$ & (0.22, 0.78) & (0.415, 0.585) \\
$\langle${overdo \latin{exaggerate}}$\rangle$ & (0.556, 0.444) & (0.319, 0.681) \\
$\langle${\black{inspect} \latin{audit} \latin{scrutinize} }$\rangle$ & (0.667, 0.25, 0.083) & (0.446, 0.429, 0.125) \\
$\langle${\germanic{sharp} \latin{acute}}$\rangle$ & (0.886, 0.114) & (0.717, 0.283) \\
$\langle${\germanic{steady} \germanic{stiff} \germanic{unwavering} \latin{firm}}$\rangle$ & (0.364, 0.172, 0.017, 0.447) & (0.278, 0.083, 0.007, 0.632) \\
$\langle${\black{ecstasy} \latin{rapture}}$\rangle$ & (0.593, 0.407) & (0.412, 0.588) \\
$\langle${\black{sizeable} \latin{ample}}$\rangle$ & (0.597, 0.403) & (0.429, 0.571) \\
$\langle${\germanic{scummy} \latin{abject} \latin{miserable}}$\rangle$ & (0.167, 0.028, 0.806) & (0.067, 0.053, 0.88) \\
$\langle${\germanic{drift} \black{displace}}$\rangle$ & (0.835, 0.165) & (0.734, 0.266) \\
$\langle${waive \latin{abandon} \black{forego}}$\rangle$ & (0.095, 0.845, 0.061) & (0.043, 0.899, 0.058) \\
$\langle${\germanic{weigh} \latin{consider} \black{count}}$\rangle$ & (0.028, 0.605, 0.367) & (0.024, 0.582, 0.394) \\
$\langle${\germanic{quick} \black{fast} \latin{rapid}}$\rangle$ & (0.328, 0.649, 0.024) & (0.326, 0.643, 0.031) \\
$\langle${\germanic{stumble} \black{stagger} \black{lurch}}$\rangle$ & (0.889, 0.097, 0.014) & (0.7, 0.114, 0.186) \\
$\langle${\black{omen} \latin{presage}}$\rangle$ & (1.0, 0.0) & (0.9, 0.1) \\
$\langle${\germanic{freight} \latin{cargo}}$\rangle$ & (0.215, 0.785) & (0.19, 0.81) \\
\end{tabular}
\caption{Top-20 examples of the most divergent usage patterns of synsets in texts of German vs.\ Spanish authors. Words with (recorded) Germanic origins are in \germanic{blue} \black{and words with (recorded) Latin origins are in} \latin{red}.}
\label{tbl:divergence-analysis}
\end{table*}

The \{\textnl{weariness, tiredness, fatigue}\} synset reveals the preference of Spanish native speakers for \textnl{fatigue}, whose Spanish equivalent \textnl{fatiga} resembles it to a great extent; \textnl{weariness}, however, is slightly more frequent in the texts of German speakers, potentially reflecting its Proto-Germanic \textnl{*w\=or\={\i}gaz} ancestor. An interesting phenomenon is revealed by the synset \{\textnl{conceivable, imaginable}\}: while both words have Latin origins, \textnl{imaginable} is more ubiquitous in the English language, rendering it more frequent in texts of German native speakers, compared to the more balanced choice of Spanish authors. Usage patterns in \{\textnl{overdo, exaggerate}\} and \{\textnl{inspect, audit, scrutinize}\} can be attributed to the same phenomenon, where the German equivalent for \textnl{inspect} (\textnl{inspizieren}) resembles its English counterpart despite a different etymological root.

\begin{table*}[h!]
\centering
\fontsize{11}{11}
\begin{tabular}{lp{13.5cm}}
\multicolumn{1}{c}{\textbf{L1}} & \multicolumn{1}{c}{\textbf{Sentence}} \\ \hline
French & \textnl{ I have to go to the Dr. to do a \textbf{rapid} check on my heart stability.} \\ \hline
French & \textnl{Maybe put every name through a manual \textbf{approbation} pipeline so it ensures quality.} \\ \hline
French & \textnl{Polls have shown public \textbf{approbation} for this law is somewhere between 58\% and 65\%, and it has been a strong promise during the presidential campaign.} \\ \hline
Italian & \textnl{The event was even more shocking because the \textbf{precedent} evening he wasn't sick at all.} \\
\end{tabular}
\caption{Cognate facilitation phenomena in usage examples by Reddit authors.}
\label{tbl:usage-examples}
\end{table*}

\paragraph{Usage examples} Table~\ref{tbl:usage-examples} presents example sentences written by Reddit authors with French and Italian L1s, further illustrating discrepancies in lexical choice (presumably) stemming from cognate facilitation effects. The French \textnl{rapide} is a translation equivalent of the English synset \{\textnl{rapid, quick, fast}\}, but its English \textnl{rapid} cognate is more constrained to contexts of movement or growth, rendering the collocation \textnl{rapid check} somewhat marked. The French noun {\textnl{approbation} is more frequent in contemporary French than its English (practically unused) equivalent \textnl{approbation}; this makes its use in English sound unnatural. In our Reddit corpus, \textnl{approbation} appears 48 times in L1-French texts, compared to 5, 4, and 4 in equal-sized texts by authors from the UK, Ireland and Canada, respectively. One of the frequent English synonym alternatives \{\textnl{approval, acceptance}\} would better fit this context. Finally, while the Italian expression \textnl{sera precedente} is common, its English equivalent \textnl{precedent evening} is very infrequent, yet it is used in English productions of Italian speakers.

\section{Conclusion}
We presented an investigation of L1 cognate effects on the productions of advanced non-native Reddit authors. The results are accompanied by a large dataset of native and non-native English speakers, annotated for author country (and, presumably, also L1) at the sentence level.

Several open questions remain for future research. From a theoretical perspective, we would like to extend this work by studying whether the tendency to choose an English cognate is more powerful in L1s with both phonetic and orthographic similarity to English (Roman script) than in L1s with phonetic similarity only (e.g., Cyrillic script). We also plan to more carefully investigate productions of speakers from multilingual countries, like Belgium and Switzerland. Another extension of this work may broaden the analysis to include additional language families.

There are also various potential practical applications to this work. First, we plan to exploit the potential benefits of our findings to the task of native language identification of (highly advanced) non-native authors, in various domains. Second, our results will be instrumental for personalization of language learning applications, based on the L1 background of the learner. For example, error correction systems can be enhanced with the native language of the author to offer root cause analysis of subtle discrepancies in the usage of lexical items, considering both their frequencies and context. Given the L1 of the target audience, lexical simplification systems can also benefit from cognate cues, e.g., by providing an informed choice of potentially challenging candidates for substitution with a simplified alternative. We leave such applications for future research.

\section*{Acknowledgments}
This work was partially supported by the National Science Foundation through award IIS-1526745.
We would like to thank Anat Prior and Steffen Eger for valuable suggestions. We are also grateful to Sivan Rabinovich for much advise and helpful comments, and to three anonymous reviewers for their constructive feedback.

%\clearpage
\vspace{0.5cm}
\small{
\bibliography{all}
\bibliographystyle{plainnat}
}

%\clearpage
%\input{appendices}

\end{document}